# Neural Copula: A unified framework for estimating generic high-dimensional Copula functions

Zhi Zeng[1,2,a], Ting Wang[1,b]

[a] School of Mechano-Electronic Engineering, Xidian University, 2 South Taibai Road, Xian, Shaan Xi, China

[b] School of Optoelectronic Engineering, Xidian University, 2 South Taibai Road, Xian, Shaan Xi, China

## Abstract

The Copula is widely used to describe the relationship between the marginal distribution and joint distribution of random variables. The estimation of high-dimensional Copula is difficult, and most existing solutions rely either on simplified assumptions or on complicating recursive decompositions. Therefore, people still hope to obtain a generic Copula estimation method with both universality and simplicity. To reach this goal, a novel neural network-based method (named Neural Copula) is proposed in this paper. In this method, a hierarchical unsupervised neural network is constructed to estimate the marginal distribution function and the Copula function by solving differential equations. In the training program, constraints are imposed on both the neural network and its derivatives. The Copula estimated by the proposed method is smooth and has an analytic expression. The effectiveness of the proposed method is evaluated on both real-world datasets and complex numerical simulations. Experimental results show that Neural Copula's fitting quality for complex distributions is significantly better than classical methods. The relevant code for the experiments is available on **GitHub**[3]. (We encourage the reader to run the program for a better understanding of the proposed method).

**Keywords**: Copula, Neural network, Probability density function, Cumulative distribution function,

---

[1] The authors have contributed equally to the paper. (Email: zhizeng@mail.xidian.edu.cn; tingw2016hg@126.com)
[2] ✉ Corresponding author.
[3] https://github.com/zengzhi2015/Neural-Copula

Neural Copula

# 1. Introduction

The Copula function is a general-purpose tool for capturing correlations between random variables, and it constructs a bridge between the joint distribution function and the marginal distribution function [1]. Sklar's theorem shows that the joint distribution of the random variables after transforming by the marginal distribution function can be represented by a Copula function [2]. Since the Copula function can accurately model complex correlations, it is widely used in financial risk management [3], signal processing [4], healthcare [5], healthy diet [6], classification problems [7], image processing [8], etc. [9][10][11].

The estimation of high-dimensional Copula is difficult [12], and most existing solutions rely either on simplified assumptions or on complicating recursive decompositions (such as the vine Copula). When one chooses among parametric [13], semiparametric [14], and nonparametric methods [15][16], a balance must be struck between accuracy and computational complexity. Most Copulas are estimated via maximum likelihood estimation (MLE) [17]. The accuracy relies heavily on the guesses of the Copula family. Once the family selection is wrong it can cause serious estimation errors. For example, in the years before 2008, the collateralized debt obligations (CDO) market relied almost exclusively on Gaussian Copula, a correlation model, for risk management. Gaussian Copula assumes that the correlation is a constant, not a variable. However, the correlation between financial data is known to be unstable. Therefore, the over-reliance on Gaussian Copula leads to the 2008-2009 Global Financial Crisis [18]. Therefore, people have always hoped to obtain a Copula estimation method with both universality and accuracy. In recent years, some scholars have tried to use deep neural networks to approximate the Archimedes Copula function and obtained valuable results [19]. This convinces us that, using sophisticated neural networks, it is possible to fit more general Copula structures with sufficient accuracy. To the best of our knowledge, there is no neural network-based method to estimate the general high-dimensional Copula function. The essential problem that needs to be solved is how to incorporate the complex constraints as loss functions into the neural network so that convergence can be easily achieved.

To reach this goal, a novel neural network-based method (named Neural Copula) is proposed in this paper. In this method, a hierarchical unsupervised neural network is constructed to estimate the cumulative distribution function (CDF) and the Copula. In the training process, constraints are imposed on both the neural network and its derivatives. The Copula estimated by the proposed method is smooth and has an analytic expression. The effectiveness of the proposed method is evaluated on both real-world datasets and complex numerical simulations.

The rest of the paper is organized as follows. Section II introduces the definition of Copula and Sklar's theorem. Section III introduces the proposed Neural Copula. Numerical experiments are given in Section IV. Section V concludes the paper.

## 2. The definition of Copula and Sklar's Theorem

This section provides a brief introduction to the definition of Copula Function and Sklar's Theorem.

**Definition 1 (Copula):**

A $d$-dimensional function $C:[0,1]^d \to [0,1]$ is called Copula, when it satisfies the following conditions, and constitutes the CDF whose marginal distribution is uniform [17][20]:

1) If $\exists j \leq d$, makes $u_j = 0$, then there is $C(\mathbf{u}) = 0$;

2) If $\exists j \leq d$, makes $\forall i \leq d$ and $i \neq j$, then $u_i = 1$ and $C(\mathbf{u}) = u_j$;

3) $\forall \mathbf{u}, \mathbf{v} \in [0,1]^d$, if $\forall i \leq d$ and $u_i < v_i$, then $\sum_{\mathbf{w} \in \times_{i=1}^d \{u_i, v_i\}} (-1)^{|i: w_i = u_i|} C(\mathbf{w}) \geq 0$.

According to this definition of Copula, the CDF of any high-dimensional random variable with known marginal distribution can be obtained, which is the famous Sklar's theorem [20].

**Theorem 1 (Sklar's Theorem):**

Suppose there is a $d$-dimensional CDF $F(\mathbf{x})$, and its marginal distributions are denoted as $\{F_1(x), F_2(x), \cdots, F_d(x)\}$. Then, there exists the Copula function that can describe the relationship between the two.

$$\begin{aligned} F(\mathbf{x}) &= C(\mathbf{u}) \\ &= C(u_1, u_2, \cdots, u_d) \\ &= C(F_1(x_1), F_2(x_2), \cdots, F_d(x_d)) \end{aligned} \quad (1)$$

If both $F(\mathbf{x})$ and $C(\mathbf{u})$ are smoothly derivable of infinite order, the probability density function (PDF) can be expressed as

$$f(\mathbf{x}) = c(\mathbf{u}) \prod_{i=1}^d f_i(x_i) \quad (2)$$

where

$$c(\mathbf{u}) = \frac{\partial^d C}{\partial u_1 \cdots \partial u_d} \quad (3)$$

## 3. Neural Copula

This section introduces the proposed Neural Copula and shows how to incorporate the complex Copula constraints into the neural network as loss functions that easily lead to convergence. The estimation of Copula is divided into two steps. First, a bunch of fully connected neural networks, named the Marginal model, are used to estimate the CDF. Then, another neural network, named the Copula model, is proposed to approximate the Copula by solving a differential equation.

For the first task, in 2019, Chi-Hua Chen et al. proposed a deep neural model which can be used to estimate the required CDF of the marginal models [21]. Other alternative and improved techniques can also be found [22][23]. The main structural difference between the proposed method and the above candidates is that we use a more compact neural representation combined with constraints imposed on both the neural network and its derivatives.

For the second task, unlike the case encountered with the Marginal Model, we know from the unique solution conditions of the partial differential equation that, it is not possible to uniquely determine the shape of the CDF if only the distribution of its derivatives and the boundary values of the CDF in higher dimensions. Thus, we first propose an observation added loss to solve this problem.

The reason we did not chose to model the PDF directly, but the CDF is that the neural network can quickly derive the key information for computing the PDF from the CDF model through an automatic derivation mechanism. However, the neural network does not have the ability to integrate out the CDF model quickly and automatically from the PDF model. In other words, if modeling the CDF, one can get the analytic expressions of both the CDF and PDF based on the neural network. However, if only modeling the PDF, one will not be able to get a concise analytic expression for the CDF. Meanwhile, the reason that the method we proposed does not directly sample the CDF before network fitting is that, on the one hand, the sampling of the CDF often involves the calculation of a large amount of statistical information and integration. On the other hand, it was found in the experiments that estimating the PDF by directly fitting the CDF and then deriving the derivative tends to introduce a large amount of noise, which is not conducive to engineering applications.

### 3.1 The structure of the Marginal Model and Copula Model

First, we use the fully connected network Marginal Model to acquire the CDF value. The model can be expressed as:

*Input layer:*

$$\mathbf{h}_m^0 = x \in \Omega \subset \mathbb{R} \tag{4}$$

where $\Omega$ is the domain of $x$.

*Hidden layer:*

$$\mathbf{h}_m^{j+1} = \tanh\left(\mathbf{w}_m^j \mathbf{h}_m^j + \mathbf{b}_m^j\right) \quad j \in \{0,1,\cdots,l_m-1\} \tag{5}$$

*Output layer:*

$$\hat{F}_m = \mathrm{sigmoid}\left(\mathbf{w}^{l_m}\mathbf{u}^{l_m} + \mathbf{b}^{l_m}\right) \in [0,1] \tag{6}$$

Taking into account that the neural network has its corresponding parameters (network weights), we denote the output of the Marginal Model as $\hat{F}_m(x,\boldsymbol{\theta}_m)$, where $\boldsymbol{\theta}_m$ represents the set of all weights of Marginal Model, that is $\boldsymbol{\theta}_m = \{\mathbf{w}_m^j\} \cup \{\mathbf{b}_m^j\}$. In accordance with the automatic derivation mechanism of the neural network, the PDF of the marginal distribution can be obtained as follows:

$$\hat{f}_m(x,\boldsymbol{\theta}_m) = \frac{d\hat{F}_m(x,\boldsymbol{\theta}_m)}{dx} \tag{7}$$

Secondly, a fully connected neural network, the Copula Model, is proposed to estimate the Copula value. Its framework also consists of three parts:

*Input layer:*

$$\mathbf{h}_c^0 = \mathbf{u} = [u_1, u_2, \cdots, u_d]^T \in [0,1]^d \tag{8}$$

*Hidden layer:*

$$\mathbf{h}_c^{j+1} = \tanh\left(\mathbf{w}_c^j \mathbf{h}_c^j + \mathbf{b}_c^j\right) \quad j \in \{0,1,\cdots,l_c-1\} \tag{9}$$

*Output layer:*

$$\hat{C} = \mathrm{sigmoid}\left(\mathbf{w}^{l_c}\mathbf{u}^{l_c} + \mathbf{b}^{l_c}\right) \in [0,1] \tag{10}$$

where $d$ in the input layer denotes the dimension of $\mathbf{u}$, where $u_i \in [0,1]$. The Copula Model has a total of $l_c - 1$ hidden layers, where $\mathbf{w}_c^j$ and $\mathbf{b}_c^j$ represent the weight and bias of the $j^{\mathrm{th}}$ hidden layer, respectively. The output $\hat{C}(\mathbf{u})$ is an estimation of the Copula function. Since Copula Model contains parameters, we denote it as $\hat{C}(\mathbf{u},\boldsymbol{\theta}_c)$, where $\boldsymbol{\theta}_c$ represents the set of all weights of the Copula Model, i.e. $\boldsymbol{\theta}_c = \{\mathbf{w}_c^j\} \cup \{\mathbf{b}_c^j\}$. Based on the automatic derivation mechanism, the PDF can be obtained as

$$\hat{c}(\mathbf{u},\boldsymbol{\theta}_c) = \frac{\partial^d \hat{C}(\mathbf{u},\boldsymbol{\theta}_c)}{\partial u_1 \partial u_2 \cdots \partial u_d} \tag{11}$$

Hence, the PDF of the sampled data can be expressed as

$$\hat{f}(\mathbf{x},\{\boldsymbol{\theta}_m^i\},\boldsymbol{\theta}_c) = \hat{c}(\hat{F}_1(x_1,\boldsymbol{\theta}_m^1),\hat{F}_2(x_2,\boldsymbol{\theta}_m^2),\cdots,\hat{F}_d(x_d,\boldsymbol{\theta}_m^d),\boldsymbol{\theta}_c) \prod_{i=1}^{d} f_i(x_i,\boldsymbol{\theta}_m^i) \tag{12}$$

The CDF of the sampled data can be expressed as

$$\hat{F}(\mathbf{x},\{\boldsymbol{\theta}_m^i\},\boldsymbol{\theta}_c) = \hat{C}(\hat{F}_1(x_1,\boldsymbol{\theta}_m^1),\hat{F}_2(x_2,\boldsymbol{\theta}_m^2),\cdots,\hat{F}_d(x_d,\boldsymbol{\theta}_m^d),\{\boldsymbol{\theta}_m^i\},\boldsymbol{\theta}_c) \tag{13}$$

### 3.2 The loss function for the Marginal Model

The proposed Marginal Model's loss function is composed of four parts:

1) The fitness of $\hat{f}_m(x,\boldsymbol{\theta}_m)$ to the sampled data can be expressed in terms of log-loss in the maximum likelihood estimation. That is

$$L_m^1(\boldsymbol{\theta}_m) = \frac{1}{n_m^1} \sum_{i=1}^{n_m^1} \left[ \log \hat{f}_m(x_i,\boldsymbol{\theta}_m) \right] \tag{14}$$

The set of all training samples corresponding to this loss is denoted as $D_m^1 = \{x_i\}_{i=1,2,\cdots n_m^1}$. The total number of samples in the training set is $n_m^1$.

2) $\hat{f}_m(x,\boldsymbol{\theta}_m)$ must be non-negative, so the penalty must be given for the negative part of $\hat{f}_m(x,\boldsymbol{\theta}_m)$, i.e.

$$L_m^2(\boldsymbol{\theta}_m) = \int_{x_m \in \Omega_m} \mathrm{relu}(-\hat{f}_m(x,\boldsymbol{\theta}_m)) dx_m \tag{15}$$

where $\Omega_m$ is the domain of $x_m$. Practically, the estimation of $L_m^2(\boldsymbol{\theta}_m)$ can be calculated by

$$L_m^2(\boldsymbol{\theta}_m) \approx \frac{1}{n_m^2} \sum_{i=1}^{n_m^2} \mathrm{relu}(-\hat{f}_m(x_i,\boldsymbol{\theta}_m)) \tag{16}$$

The set of all training samples corresponding to this loss is denoted as $D_m^2 = \{x_i\}_{i=1,2,\cdots n_m^2}$. Without loss of generality, the positions of the samples can be uniformly distributed over the domain.

3) The integral of $\hat{f}_m(x, \boldsymbol{\theta}_m)$ over the universe must be 1, so the corresponding loss function is as follows.

$$L_m^3(\boldsymbol{\theta}_m) = \left| 1 - \int_{x_m \in \Omega_m} \hat{f}_m(x, \boldsymbol{\theta}_m) dx_m \right| \tag{17}$$

where $\Omega_m$ is the domain of $x_m$.

Practically, the estimation of $L_m^3(\boldsymbol{\theta}_m)$ can be calculated by

$$L_m^3(\boldsymbol{\theta}_m) \approx \left| 1 - \frac{1}{n_m^3} \sum_{i=1}^{n_m^3} \hat{f}_m(x_i, \boldsymbol{\theta}_m) \Delta_m \right| \tag{18}$$

The set of all training samples corresponding to this loss is denoted as $D_m^3 = \{x_i\}_{i=1,2,\cdots n_m^3}$. Without loss of generality, the positions of the samples in this dataset can be uniformly distributed over its domain, and the interval of adjacent data is denoted as $\Delta_m$.

4) The CDF $\hat{F}_m(x, \boldsymbol{\theta}_m)$ must satisfy the following conditions

$$\begin{cases} \hat{F}_m(0, \boldsymbol{\theta}_m) = 0 \\ \hat{F}_m(1, \boldsymbol{\theta}_m) = 1 \end{cases} \tag{19}$$

The corresponding loss function can be expressed as

$$L_m^4(\boldsymbol{\theta}_m) = \hat{F}_m(0, \boldsymbol{\theta}_m) + \left| 1 - \hat{F}_m(1, \boldsymbol{\theta}_m) \right| \tag{20}$$

The set of all training samples corresponding to this loss is $D_m^4 = \{0, 1\}$.

In conclusion, the loss functions of the Marginal Model can be formulated by a linear combination as

$$L_m(\boldsymbol{\theta}_m) = \sum_{k=1}^{4} \lambda_k L_m^k(\boldsymbol{\theta}_m) \tag{21}$$

**3.3 The loss function for the Copula Model**

Similar to the Marginal Model, the constraint (loss function) acting on the Copula Model consists of five components:

1) The fitness of $\hat{f}_c\left(\mathbf{x},\{\boldsymbol{\theta}_m^i\},\boldsymbol{\theta}_c\right)$ to the data distribution can be represented by the log loss in the maximum likelihood estimation. That is

$$L_c^1\left(\{\boldsymbol{\theta}_m^i\},\boldsymbol{\theta}_c\right) = \frac{1}{n_c^1}\sum_{i=1}^{n_c^1}\left[\log \hat{f}_c\left(\mathbf{x}_i,\{\boldsymbol{\theta}_m^i\},\boldsymbol{\theta}_c\right)\right] \tag{22}$$

We denote the set of all training samples corresponding to this loss as $D_c^1 = \{\mathbf{x}_i\}_{i=1,2,\cdots n_c^1}$.

2) $\hat{f}_c\left(\mathbf{x},\{\boldsymbol{\theta}_m^i\},\boldsymbol{\theta}_c\right)$ must be non-negative, thus the penalty must be given for the negative part of $\hat{f}_c\left(\mathbf{x},\{\boldsymbol{\theta}_m^i\},\boldsymbol{\theta}_c\right)$, i.e

$$L_c^2\left(\{\boldsymbol{\theta}_m^i\},\boldsymbol{\theta}_c\right) = \int_{x_1\in\Omega_1}\int_{x_2\in\Omega_2}\cdots\int_{x_d\in\Omega_d} \mathrm{relu}\left(-\hat{f}_c\left(\mathbf{x}_i,\{\boldsymbol{\theta}_m^i\},\boldsymbol{\theta}_c\right)\right)dx_1 dx_2 \cdots dx_d \tag{23}$$

To simplify the calculation, we estimate the numerical approximation of the above loss function by

$$L_c^2\left(\{\boldsymbol{\theta}_m^i\},\boldsymbol{\theta}_c\right) \approx \frac{1}{n_c^2}\sum_{i=1}^{n_c^2}\mathrm{relu}\left(-\hat{f}_c\left(\mathbf{x}_i,\{\boldsymbol{\theta}_m^i\},\boldsymbol{\theta}_c\right)\right) \tag{24}$$

The set of all training samples corresponding to this loss is denoted as $D_c^2 = \{\mathbf{x}_i\}_{i=1,2,\cdots n_c^2}$. Without loss of generality, the positions of the samples of this dataset are uniformly distributed over its domain.

3) The integration of $\hat{f}_c\left(\mathbf{x},\{\boldsymbol{\theta}_m^i\},\boldsymbol{\theta}_c\right)$ over the universe must be 1, so the corresponding loss function is defined as follows:

$$L_c^3\left(\{\boldsymbol{\theta}_m^i\},\boldsymbol{\theta}_c\right) = \left|1 - \int_{x_1\in\Omega_1}\int_{x_2\in\Omega_2}\cdots\int_{x_d\in\Omega_d} \hat{f}_c\left(\mathbf{x}_i,\{\boldsymbol{\theta}_m^i\},\boldsymbol{\theta}_c\right)dx_1 dx_2 \cdots dx_d\right| \tag{25}$$

Practically, the estimation of $L_c^3\left(\{\boldsymbol{\theta}_m^i\},\boldsymbol{\theta}_c\right)$ can be calculated by

$$L_c^3\left(\{\boldsymbol{\theta}_m^i\},\boldsymbol{\theta}_c\right) \approx \left|1 - \frac{1}{n_c^3}\sum_{i=1}^{n_c^3}\hat{f}_c\left(\mathbf{x}_i,\{\boldsymbol{\theta}_m^i\},\boldsymbol{\theta}_c\right)\prod_{j=1}^{d}\Delta_j\right| \tag{26}$$

The set of all training samples corresponding to this loss is denoted as $D_c^3 = \{\mathbf{x}_i\}_{i=1,2,\cdots n_c^3}$. The positions of the samples of this dataset are uniformly distributed over its domain, and the interval of adjacent data along the $j^{th}$ dimension is denoted as $\Delta_j$.

4) The distribution of $\hat{C}(\mathbf{u}, \boldsymbol{\theta}_c)$ on the boundary must satisfy Copula's definition, that is

$$\begin{cases} \hat{C}(\underline{\mathbf{u}}_1, \boldsymbol{\theta}_c) = \hat{C}(0, u_2, \cdots, u_d, \boldsymbol{\theta}_c) = 0 \\ \hat{C}(\underline{\mathbf{u}}_2, \boldsymbol{\theta}_c) = \hat{C}(u_1, 0, \cdots, u_d, \boldsymbol{\theta}_c) = 0 \\ \vdots \\ \hat{C}(\underline{\mathbf{u}}_d, \boldsymbol{\theta}_c) = \hat{C}(u_1, u_2, \cdots, 0, \boldsymbol{\theta}_c) = 0 \end{cases}, \quad u_i \in [0,1] \tag{27}$$

$$\begin{cases} \hat{C}(\overline{\mathbf{u}}_1, \boldsymbol{\theta}_c) = \hat{C}(u_1, 1, \cdots, 1, \boldsymbol{\theta}_c) = u_1 \\ \hat{C}(\overline{\mathbf{u}}_2, \boldsymbol{\theta}_c) = \hat{C}(1, u_2, \cdots, 1, \boldsymbol{\theta}_c) = u_2 \\ \vdots \\ \hat{C}(\overline{\mathbf{u}}_d, \boldsymbol{\theta}_c) = \hat{C}(1, 1, \cdots, u_d, \boldsymbol{\theta}_c) = u_d \end{cases}, \quad u_i \in [0,1] \tag{28}$$

The corresponding loss function can be stated as

$$L_c^4\left(\{\boldsymbol{\theta}_m^i\}, \boldsymbol{\theta}_c\right) = \sum_{i=1}^{d} \sum_{j=1}^{n_c^4} \hat{C}\left(\underline{\mathbf{u}}_i^j, \boldsymbol{\theta}_c\right) + \sum_{i=1}^{d} \sum_{j=1}^{n_c^4} \left|\hat{C}\left(\overline{\mathbf{u}}_i^j, \boldsymbol{\theta}_c\right) - u_i^j\right| \tag{29}$$

We denote the set of all training samples corresponding to this loss as $D_c^4 = \bigcup_{i=1}^{d}\left(\{\underline{\mathbf{u}}_i^j\}_{j=1,2,\cdots n_c^4} \cup \{\overline{\mathbf{u}}_i^j\}_{j=1,2,\cdots n_c^4}\right)$. Without loss of generality, the positions of the samples of this dataset are randomly distributed in the boundary region.

5) Unlike the case encountered with the Marginal Model, we know from the unique solution conditions of the partial differential equation that, it is not possible to uniquely determine the shape of the CDF if only the distribution of its derivatives and the boundary values of the CDF in higher dimensions ($d \geq 3$) are given. For this reason, certain observation values of the CDF must be added to ensure the uniqueness of the estimation. It should be noted that it is usually time-consuming to sample from the CDF, so it is not recommended to calculate an excessive number of observations and then fit it through the network. The values of $\hat{F}_c\left(\mathbf{x}, \{\boldsymbol{\theta}_m^i\}, \boldsymbol{\theta}_c\right)$ at certain observation points are consistent with the statistical law, that is

$$\hat{F}_c\left(\mathbf{x},\{\boldsymbol{\theta}_m^i\},\boldsymbol{\theta}_c\right) = \int_{-\infty}^{x_d} \cdots \int_{-\infty}^{x_2} \int_{-\infty}^{x_1} \hat{f}_c\left(\mathbf{y},\{\boldsymbol{\theta}_m^i\},\boldsymbol{\theta}_c\right) dy_1 dy_2 \cdots dy_d \tag{30}$$

Eq. (30) can be approximated by

$$\hat{F}_c\left(\mathbf{x},\{\boldsymbol{\theta}_m^i\},\boldsymbol{\theta}_c\right) \approx \frac{1}{n_c^1} \sum_{\mathbf{y} \in D_c^1} \text{flag}(\mathbf{x},\mathbf{y}) \tag{31}$$

where

$$\text{flag}(\mathbf{x},\mathbf{y}) = \begin{cases} 1 & \forall j \le d,\ y_j < x_j \\ 0 & \text{otherwise} \end{cases} \tag{32}$$

The loss function corresponding to the above constraints can be formulated as

$$L_c^5\left(\{\boldsymbol{\theta}_m^i\},\boldsymbol{\theta}_c\right) = \frac{1}{n_c^1 n_c^5} \sum_{i=1}^{n_c^5} \left| \hat{F}_c\left(\mathbf{x}_i,\{\boldsymbol{\theta}_m^i\},\boldsymbol{\theta}_c\right) - \sum_{\mathbf{y} \in D_c^1} \text{flag}(\mathbf{x}_i,\mathbf{y}) \right| \tag{33}$$

We denote the set of all training samples corresponding to this loss as $D_c^5 = \{\mathbf{x}_i\}_{i=1,2,\cdots n_c^5}$. Without loss of generality, the samples of this dataset are randomly distributed in the domain.

Finally, we combine all the above loss functions of the Copula Model using the linear combination. That is

$$L_c\left(\{\boldsymbol{\theta}_m^i\},\boldsymbol{\theta}_c\right) = \sum_{k=1}^{5} \lambda_k L_k\left(\{\boldsymbol{\theta}_m^i\},\boldsymbol{\theta}_c\right) \tag{34}$$

### 3.4 Summary

This section summarizes Neural Copula's neural network architecture and its training framework in several schematic images.

The framework of the Neural Copula is shown in Fig. 1, which consists of several Marginal Models (the number of which depends on the dimensions of the input data) and a Copula Model. Each channel (dimension) of the input data is fed into its corresponding Marginal Model to estimate the CDF value, and then all CDF values are fed into the Copula Model to obtain the Copula value.

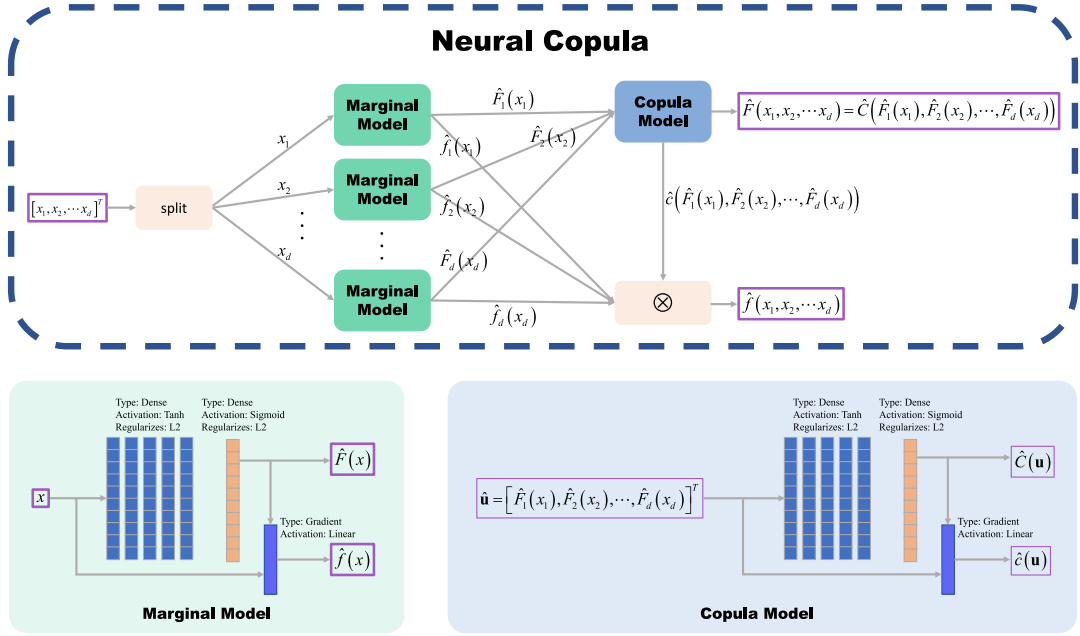

**Figure 1.** The framework for Neural Copula and its submodules: the Marginal Model and Copula Model. The elements in the purple boxes indicate the input and output data. 'split' means the samples are separated dimension by dimension. ⊗ represents the multiplication operation. The gray arrows indicate the data flow.

As illustrated in Fig. 2, (A) and (B) are the training schemes for the Marginal Model and Copula Model, respectively. Neural Copula can be trained using an end-to-end way. The Marginal Model is trained by feeding four training sets in parallel and optimizing the parameters via back-propagation. The Copula model is trained by feeding five training sets in parallel and optimizing the parameters via back-propagation. As shown in Fig. 2 (B), when updating the Copula Model, the parameters of the trained Marginal Models are fixed.

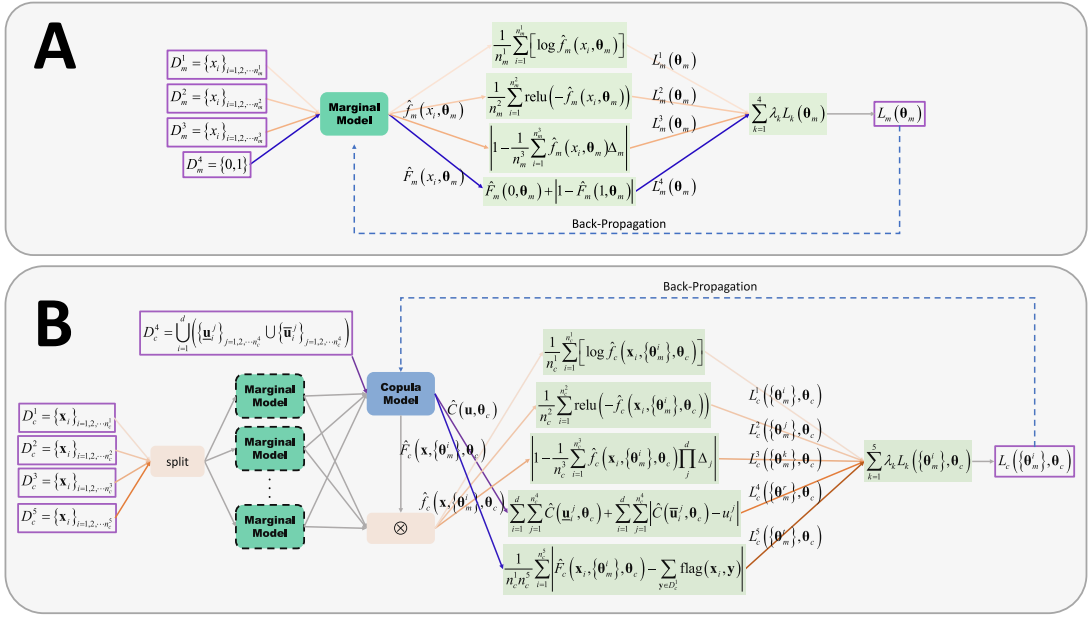

**Figure 2.** The training schemes for the Marginal Model (A) and Copula Model (B). The elements in the purple boxes indicate the input and output data. The contents in the light green box are the loss functions. The colored arrows denote that the data (marked on the arrows) should be input in parallel. The gray arrows indicate the data flow. The blue dashed arrows indicate back propagation. The black dashed box of the Marginal Model in (B) indicates that the weights of the Marginal Model are fixed when the Copula Model is training. 'split' means the samples are separated dimension by dimension. $\otimes$ represents the multiplication operation.

Figure 3 illustrates the training results of a Neural Copula for fitting samples generated by a two-dimensional Gaussian function. The estimated marginal distributions by Marginal Models and estimated CDF and PDF values by the Neural Copula are given. The generated samples according to the estimated PDF are also shown, which is to demonstrate more intuitively the accuracy of the proposed method.

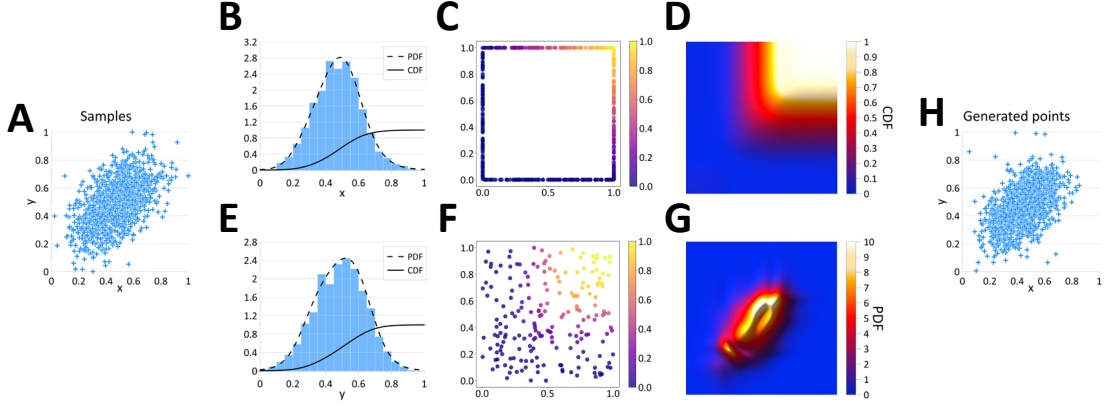

**Figure 3.** The training results of the Neural Copula for fitting samples generated by a two-dimensional Gaussian function. (A) Scatterplot of the samples. (B) and (E) are the marginal distributions. (C) and (F) are sampled boundary points and observation points, respectively. (D) and (G) are the estimated CDF and PDF by Neural Copula, respectively. (H) is the scatterplot of generated samples from Neural Copula.

## 4. Experiments

With no loss of generality, this chapter demonstrates the effectiveness of Neural Copula through 2D and 3D examples. Although our proposed method can also be used for higher-dimensional data, it will make the comparison and visualization of algorithm performance more difficult. For this reason, complex 2D and 3D cases are deliberately chosen for this experiment to show the generalizability, robustness, and accuracy of the algorithm.

### 4.1 Real-world Datasets

To evaluate the effectiveness of Neural Copula in estimating generic Copula functions, the performance of Neural Copula has been compared with three classical parametric Copula models on three real-world datasets.

The Real-world dataset used in this paper is the same as [19]. Three real-world datasets are Boston Housing, INTC-MSFT, and GOOG-FB. The data in these datasets have a limited distribution area, and the data in two dimensions are correlated. This real-world data generally has only a small number of outliers far from the distribution. Before training the network, the data in each dimension is Min-Max normalized to the range [0, 1]. This ensures that the empirical marginals are approximately uniform [19]. For real-world data, the overfitting is mild or even negligible. Therefore, the data is only divided into training and test sets, which account for 2/3 and 1/3 of the total data, respectively.

In this experiment, the sample numbers of each training set are as follows: the total number of samples in the training set is $n_m^1$, $n_m^2 = n_m^3 = 1000$. $n_c^1$ is equal to the total number of samples in the training set, $n_c^2 = n_c^3 = 2500$, $n_c^4 = 800$, $n_c^5 = 200$. The Marginal Model for the estimation of the CDF values is structured by a 1-5-5-5-5-5-1 fully connected network with an L2 regularization rate of 0.001 for each layer. Other settings include $\lambda_1 = 0.1$, $\lambda_2 = 1$, $\lambda_3 = 2$, $\lambda_4 = 2$. The Adam optimizer is used with a learning rate of 0.001. The Copula Model is structured by using a 2-10-10-10-10-10-1 fully connected network with an L2 regularization rate of 0.001 for each layer. Other settings include $\lambda_1 = 0.1$, $\lambda_2 = 1$, $\lambda_3 = 1$, $\lambda_4 = 2$, $\lambda_5 = 5$. The Adam optimizer is used with a learning rate of 0.0001. The following describes each of the three datasets used in this paper.

**Boston Housing**: In the Boston Housing dataset, there is a negative correlation between the percentage of the population with lower status and the median value of owner-occupied homes, but they are not linearly related. This dataset is from Kaggle and has 506 samples.

**INTC-MSFT**: The daily opening prices of Intel Corporation (INTC) and Microsoft (MSFT) have an approximate positive correlation. Data close to the origin of the axes show a strong positive correlation. Data far from the origin of the axes are weakly correlated, and they are more scattered with a distinct bifurcation. The INTC-MSFT dataset is derived from Yahoo Finance and has a total of 1263 samples. The INTC-MSFT dataset is the daily opening prices for INTC and MSFT from

1996 to 2000.

**GOOG-FB**: There is a strong positive correlation between the daily opening prices of Alphabet Inc. (GOOG) and Facebook (FB). The GOOG_FB dataset has a dense data distribution and two distinct clustering centers. This dataset is collected by Yahoo Finance and contains 1259 daily closing prices from May 2015 to May 2020.

We compared the performance of Neural Copula with t-Copula as the one used in [24], Frank Copula as the one used in [25], and Gaussian Copula as the one used in [26]. These three models are the most classical methods for estimating Copula and are widely used in many fields such as finance and signal processing. The performance of the four methods is measured through qualitative and quantitative comparisons. The estimation accuracy of the models is compared qualitatively based on how well the marginal and joint distributions fit the shape of the distribution of the samples; The quantitative analysis reports the log-loss of Neural Copula and the three classical Copula models on the testing dataset. The above comparisons reveal that Neural Copula can obtain an estimation accuracy no less than that of three classical methods.

### 4.1.1 Experiments on the Boston Housing dataset

For the Boston Housing dataset, as shown in the first row of Fig. 4, Neural Copula can obtain marginal distribution comparable to the three classical Copula models. As shown in the second row of Fig. 4, the joint distribution yielded by Neural Copula is also close to that of the three classical Copula models. Visually, the estimations from Neural Copula seem more similar to the data distribution as shown in Fig. 4 (A). It can be seen from Table 1 that the log loss of Neural Copula on the test set is higher than that of the other three classical Copula models. (The larger the log-loss is, the more accurate the estimation is.)

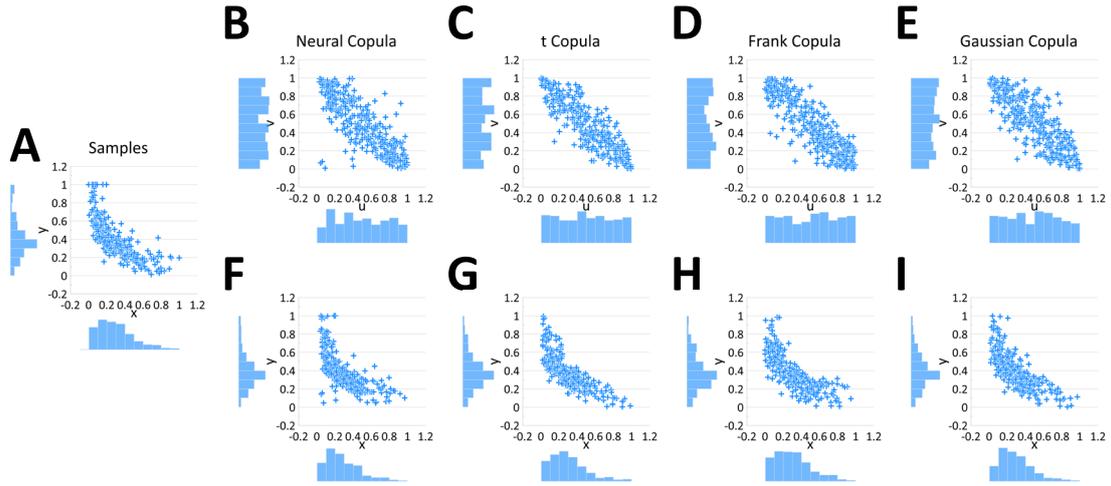

**Figure 4.** Experiment results on the Boston Housing dataset. (A) Scatter plot of the data on the training set. (B-E) The marginal distribution of Neural Copula, t-Copula, Frank Copula, and Gaussian Copula. (F-I) Scatterplot generated from PDF values of the joint distribution obtained by Neural Copula, t-Copula, Frank Copula and Gaussian Copula.

**Table 1.** Test log-loss of Neural Copula, t-Copula, Frank Copula, and Gaussian Copula on the Boston Housing dataset.

| Neural Copula | t-Copula | Frank Copula | Gaussian Copula |
| --- | --- | --- | --- |
| **1.269** | 1.242 | 1.251 | 1.229 |

### 4.1.2 Experiments on the INTC-MSFT dataset

For the INTC-MSFT dataset, as shown in the second row of Fig. 5, the joint distribution of Neural Copula is more similar to Fig. 5 (A) and much better than the three classical models. The marginal distribution of the joint distribution of Neural Copula is also closer to the original distribution than the three classical methods. For the INTC-MSFT dataset, compared with the three classical Coupla models, Neural Copula is able to model not only the data with higher dependence in the bottom left but also the data with lower dependence in the upper right. As can be seen from Table 2, the log loss of Neural Copula on the test set is considerably higher than that of the three classical Coupla models, which quantitatively indicates that Neural Copula has high estimation accuracy.

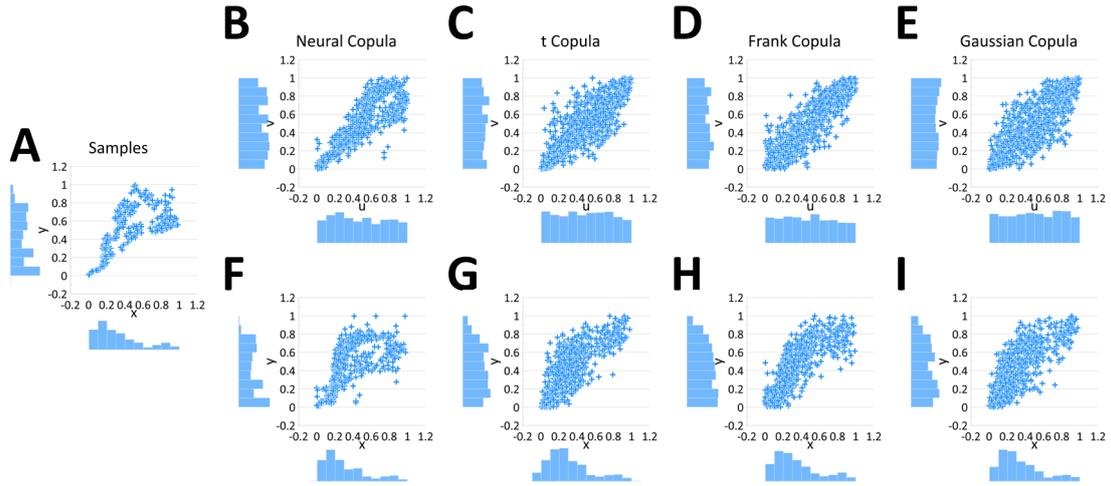

**Figure 5.** Experiment results on the INTC-MSFT dataset. (A) Scatter plot of the data on the training set. (B-E) The marginal distribution of Neural Copula, t-Copula, Frank Copula, and Gaussian Copula. (F-I) Scatterplot generated from PDF values of the joint distribution obtained by Neural Copula, t-Copula, Frank Copula and Gaussian Copula.

**Table 2.** Test log loss of Neural Copula, t-Copula, Frank Copula, and Gaussian Copula on the INTC-MSFT dataset.

| Neural Copula | t-Copula | Frank Copula | Gaussian Copula |
| --- | --- | --- | --- |
| **1.729** | 0.917 | 0.992 | 0.871 |

### 4.1.3 Experiments on the GOOG-FB dataset

For the GOOG-FB dataset, as shown in the second row of Fig. 6, the joint distribution of Neural Copula is far better than that of the three classical models, and more similar to Fig. 6 (A). Meanwhile, the marginal distribution of the joint distribution of Neural Copula is closer to the original distribution in Fig. 6 (A) than the three classical methods. As shown in the second row of Fig. 6, Neural Copula has an exceptional ability to learn the characteristics that the original distribution has two clustering centers, while the joint distribution of the three classical Copula models only has one clustering center. In Table 2, the log loss of Neural Copula on the test set is significantly higher than that of the three classical Copula models.

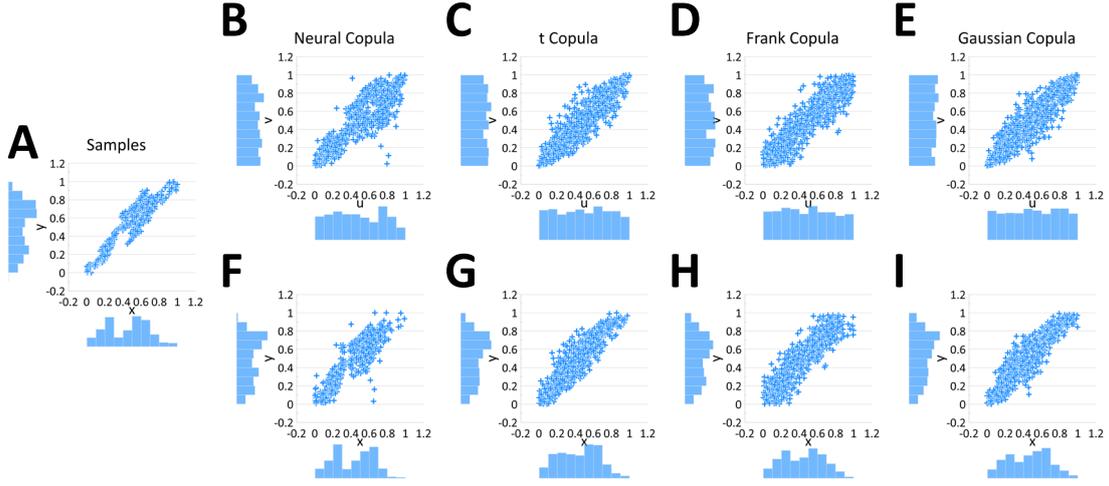

**Figure 6.** Experiment results on the GOOG-FB dataset. (A) Scatter plot of the data on the training set. (B-E) The marginal distribution of Neural Copula, t-Copula, Frank Copula, and Gaussian Copula. (F-I) Scatterplot generated from PDF values of the joint distribution obtained by Neural Copula, t-Copula, Frank Copula and Gaussian Copula.

**Table 3.** Test log-loss of Neural Copula, t-Copula, Frank Copula, and Gaussian Copula on the GOOG-FB dataset.

| Neural Copula | t-Copula | Frank Copula | Gaussian Copula |
|---|---|---|---|
| **1.509** | 1.151 | 1.056 | 1.144 |

In summary, for the above three real-world datasets, Neural Copula qualitatively and quantitatively outperforms t-Copula, Frank Copula, and Gaussian Copula.

The CPU of the computing platform used in this paper is Intel Core i9 9900X, the GPU is NVIDIA Titan RTX, and the memory is 128 GB. In terms of computational efficiency, Neural Copula takes approximately 290s to train 60,000 epochs, which can be significantly reduced if transfer learning is used, which is suitable for some applications with higher demand for better accuracy but less demanding real-time requirements.

**4.2 Complex simulation datasets**

For many applications, the distribution of the data is complex and has multiple clustering centers, which is not easy or even impossible to estimate by traditional Copula models. Therefore, to further evaluate the performance of the Neural Copula model, it is tested on two complex simulation datasets. A comparison of the performance of the Neural Copula with the other three classical parametric Copula models shows that the Neural Copula has a significant advantage in terms of fitting accuracy.

The following 2D and 3D complex simulation datasets are generated by known PDF, and consist of 2601 and 9261 samples, respectively. The distributions of the two datasets contain distinct bar-shaped (or block-shaped) clusters. Furthermore, they are far more complicated than those in previous experiments.

### 4.2.1 Two-dimensional problem

The 2601 samples in the 2D complex simulation dataset are generated randomly by the following PDF.

$$f(x,y) = \frac{1}{a}\left[\sin(x+ky)+1\right] \quad x \in [0,3] \quad y \in [0,2] \tag{35}$$

where $k = 5$, $a = 5.835386372$.

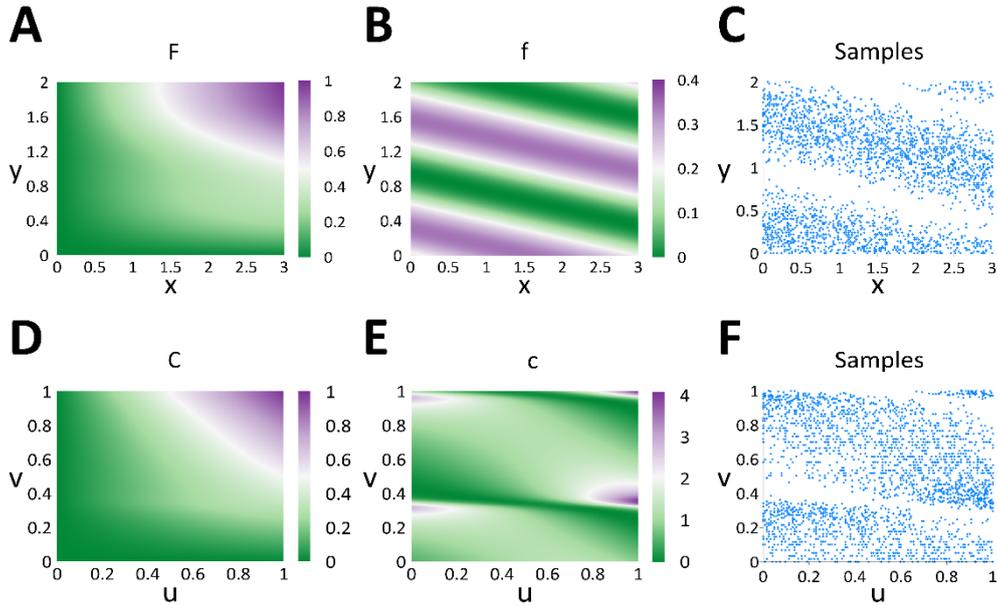

**Figure 7.** Two-dimensional Copula to be estimated. A. The CDF $F(x,y)$; B. The PDF $f(x,y)$; C. Sampling results according to $f(x,y)$; D. Copula function $C(u,v)$; E. The PDF $c(u,v)$; F. Sampling results according to $c(u,v)$.

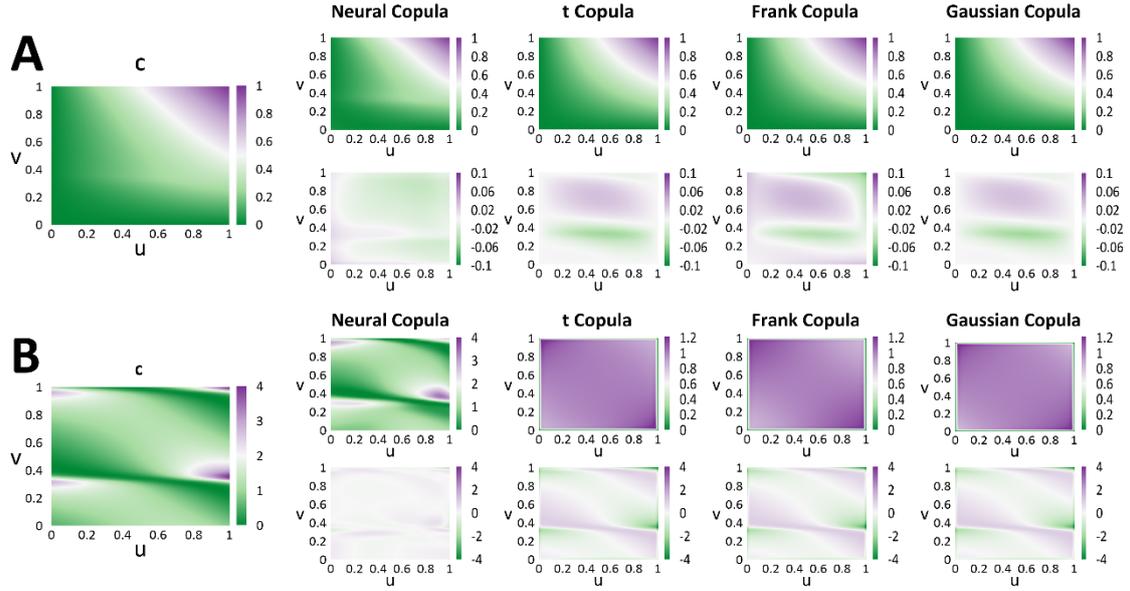

**Figure 8.** Two-dimensional Copula estimation comparisons. A. Comparison of Copula $C(u,v)$ estimation results; B. Comparison of estimation results of PDF $c(u,v)$. Large images on the left are the ground truths. The first row on the right shows the Copula estimation result; the second row shows the Copula estimation error; the third row shows the PDF estimation results; the fourth row shows the PDF estimation error. The second to fifth columns in the figure show the estimation results by Neural Copula, t-Copula, Frank-Copula, and Gaussian Copula, respectively.

The training results of the Marginal Model and Copula Model are shown in Fig. 8 (A) and Fig. 8 (B), respectively.

From the first subplot on the left of Fig. 8 (B), it can be seen that the probability distribution of the Copula function has two obvious bar-shaped clusters, which makes it difficult to fit through three classical models. As can be seen in Fig 8 (B), these three classical models incorrectly capture the pattern of the distribution of the data on the u-v space. Consequently, the fitting error is relatively large. In contrast, Neural Copula is able to accurately approximate the distribution of Copula due to the precise constraints put on its derivatives. As can be seen from the error distribution in Fig 8, the average fitting errors of Neural Copula for the Copula values and the PDF values of the Copula function are no more than 0.012 and 0.113, respectively, which can satisfy most engineering applications with less demanding requirements for the real-time performance.

### 4.2.2 Three-dimensional problem

The 9261 samples in the 3D complex simulation dataset are generated randomly by the following PDF.

$$f(x,y,z) = \frac{4.843}{\sin(x+ky+lz)+1} \qquad x \in [0,3] \; y \in [0,2] \tag{36}$$

where $k = 2$, $l = 3$.

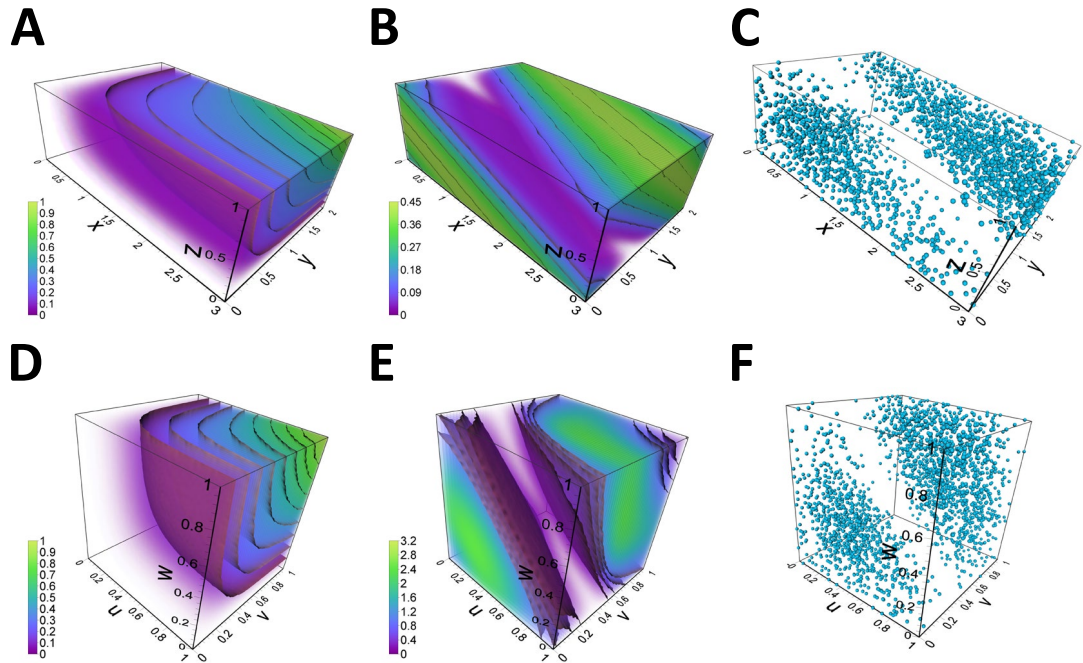

**Figure 9.** Three-dimensional Copula to be estimated. A. The CDF $F(x,y,z)$; B. The PDF $f(x,y,z)$; C. Sampling results according to $f(x,y,z)$; D. Copula $C(u,v,w)$; E. The PDF $c(u,v,w)$; F. Sampling results according to $c(u,v,w)$.

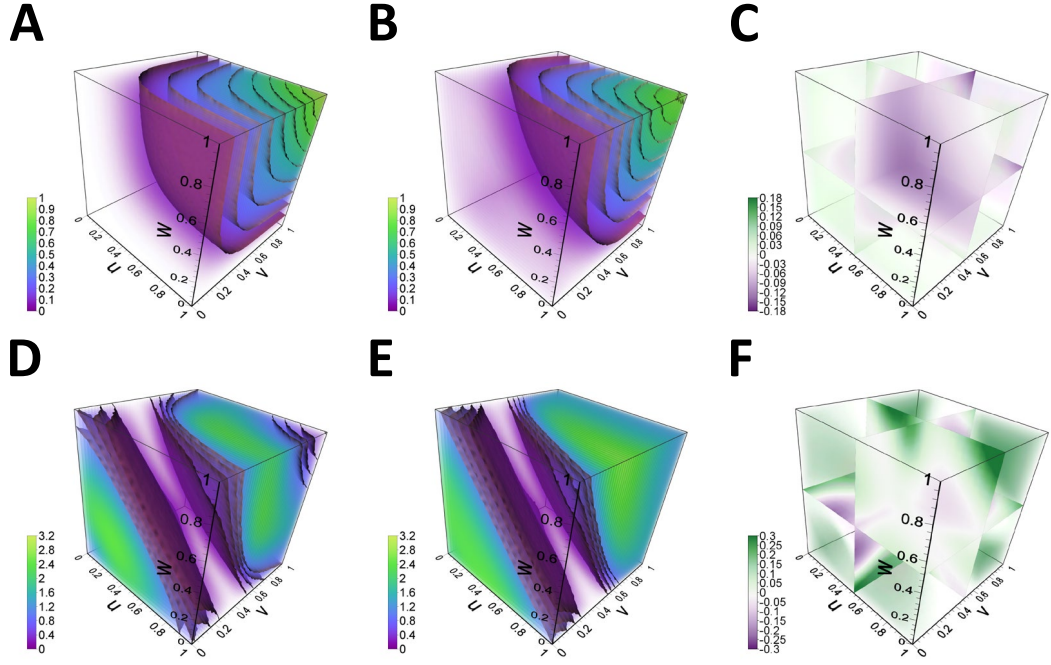

**Figure 10.** Three-dimensional Copula estimation results. A. The Copula function $C(u,v,w)$ to be estimated; B. Estimation of $C(u,v,w)$; C. Estimation error of $C(u,v,w)$; D. The distribution $c(u,v,w)$ to be estimated; E. Estimation of $c(u,v,w)$; F. Estimation error of $c(u,v,w)$.

As shown in Fig. 10 (A), the 3D samples of the probability distribution of the Copula function have two distinct clusters and a large space between the two clustering centers. Neural Copula can accurately fit the probability density function and, at the same time, can exactly approximate the Copula function. As can be seen in Fig. 10 (C) and (F), the average fitting error of Neural Copula for the Copula values and the PDF values is considerably small. These two errors are 0.039 and 0.149, respectively, which can satisfy most engineering application scenarios.

## 5. Conclusion

In this paper, we propose a neural network-based framework (Neural Copula) to estimate generic Copula functions. The method uses the Marginal Model to estimate marginal CDFs and uses the Copula Model to approximate the Copula function. The Copula function obtained by this method has good properties of analytic expression and smooth differentiability. The effectiveness of the proposed method is evaluated on both real-world datasets and complex numerical simulations.

## CRediT authorship contribution statement

**Zhi Zeng:** Conceptualization, Methodology, Formal analysis, Writing; **Ting Wang:** Writing-review & editing, Validation, Experiments.

## Declaration of Competing Interest

The authors declare that they have no known competing financial interests or personal relationships that could have appeared to influence the work reported in this paper.

## Acknowledgments

This research is supported by the financial support from the National Natural Science Foundation of China under No. 61805185.